# AN EFFICIENT CODEBOOK INITIALIZATION APPROACH FOR LBG ALGORITHM

Arup Kumar Pal[1] and Anup Sar[2]

[1]Department of Computer Science and Engineering, NIT Jamshedpur, India
arupkrpal@gmail.com
[2]Department of Electronics and Telecommunication Engineering, Jadavpur University, India
anupsareng@gmail.com

***ABSTRACT***

*In VQ based image compression technique has three major steps namely (i) Codebook Design, (ii) VQ Encoding Process and (iii) VQ Decoding Process. The performance of VQ based image compression technique depends upon the constructed codebook. A widely used technique for VQ codebook design is the Linde-Buzo-Gray (LBG) algorithm. However the performance of the standard LBG algorithm is highly dependent on the choice of the initial codebook. In this paper, we have proposed a simple and very effective approach for codebook initialization for LBG algorithm. The simulation results show that the proposed scheme is computationally efficient and gives expected performance as compared to the standard LBG algorithm.*

## 1. INTRODUCTION

Vector Quantization (VQ) [1-2] is one of the widely used and efficient techniques for image compression. VQ has received a great attention in the field of multimedia data compression [3-5] since last few decades because it has simple decoding structure and can provide high compression ratio. In order to compress the image using VQ, initially image is decomposed into non-overlapping sub image blocks, also termed as vectors. From all these blocks (or vectors), a set of representative image blocks (or vectors) are selected to represent the entire set of image blocks. The set of representative image vectors is called a codebook and each representative image vector is called codeword.

In this paper, we have employed VQ for image compression. In VQ, the most important task is designing an efficient codebook. There are already several algorithms [6-12] published on how to generate a codebook. The *LBG algorithm* [6] is the most cited and widely used algorithm on designing the VQ codebook. It is the starting point for most of the work on vector quantization. The performance of the LBG algorithm is extremely dependent on the selection of the initial codebook. In conventional LBG algorithm, the initial codebook is chosen at random from the training data set. It is observed that some-time it produces poor quality codebook. Due to the bad codebook initialization, it always converges to the nearest local minimum. This problem is called the local optimal problem [10]. In addition, it is observed that the time required to complete the iterations depends upon how good the initial codebook is. In literature [10-13], several initialization techniques have been reported for obtaining a better local minimum.

In order to alleviate problems associated with the codebook initialization for the LBG algorithm, in this paper, we have proposed a novel codebook initialization technique. In the initialization process, we have first chosen a highest level approximate image from the original image using image pyramid [14] and subsequently the selected highest level approximated image is decomposed into blocks to select as the initial codebook for codebook generation. The selected initial codebook is trained into an improve one through several iterative processes. The proposed algorithm has been implemented and tested on a set of standard test images and the performance is compared with respect to the standard *LBG* algorithm.

The rest of the paper is organized as follows. A brief overview VQ is given in section 2. The proposed scheme is elaborated in section 3. Experimental results are presented in section 4 to discuss the relative performance of the proposed scheme with the standard LBG algorithm. Finally, conclusions are given in section 5.

## 2. VECTOR QUANTIZATION

The concept of VQ is based on Shannon's rate-distortion theory where it says that the better compression is always achievable by encoding sequences of input samples rather than the input samples one by one. In VQ based image compression, initially image is decomposed into non-overlapping sub image blocks. Each sub block is then converted into one-dimension vector which is termed as training vector. From all these training vectors, a set of representative vectors are selected to represent the entire set of training vectors. The set of representative training vectors is called a codebook and each representative training vector is called codeword. Thus it can be redefined as a mapping function $Q$ from an input vector in $\square$-dimensional Euclidean space $R^{\square}$ into a finite subset $C$ of $R^{\square}$ i.e. $Q: R^{\square} \rightarrow C$, where $C=\{ c_i \mid i=1,2,\ldots,N\}$ is the codebook of size $N$ and each $c_i=(c_{i1},c_{i2},\ldots,c_{i\square})^T$ is called a codeword in the codebook $C$.

As stated earlier, the VQ process is done in the following three steps namely (i) codebook design, (ii) encoding process and (iii) decoding process. In *LBG* algorithm an initial codebook is chosen at random from the training vectors. Then the initial codebook is trained into an improved one by several iteration processes. The flowchart of *LBG* clustering algorithm is shown in Figure 1. After codebook design process, each codeword of the codebook is assigned a unique index value. Then in the encoding process, any arbitrary vector corresponding to a block from the image under consideration is replaced by the index of the most appropriate representative codeword. The matching is done based on the computation of minimum squared Euclidean distance between the input training vector and the codeword from the codebook. So after encoding process, an index-table is produced. The codebook and the index-table is nothing but the compressed form of the input image. In decoding process, the codebook which is available at the receiver end too, is employed to translate the index back to its corresponding codeword. This decoding process is simple and straight forward. Figure 2 shows the schematic diagram of VQ encoding-decoding process.

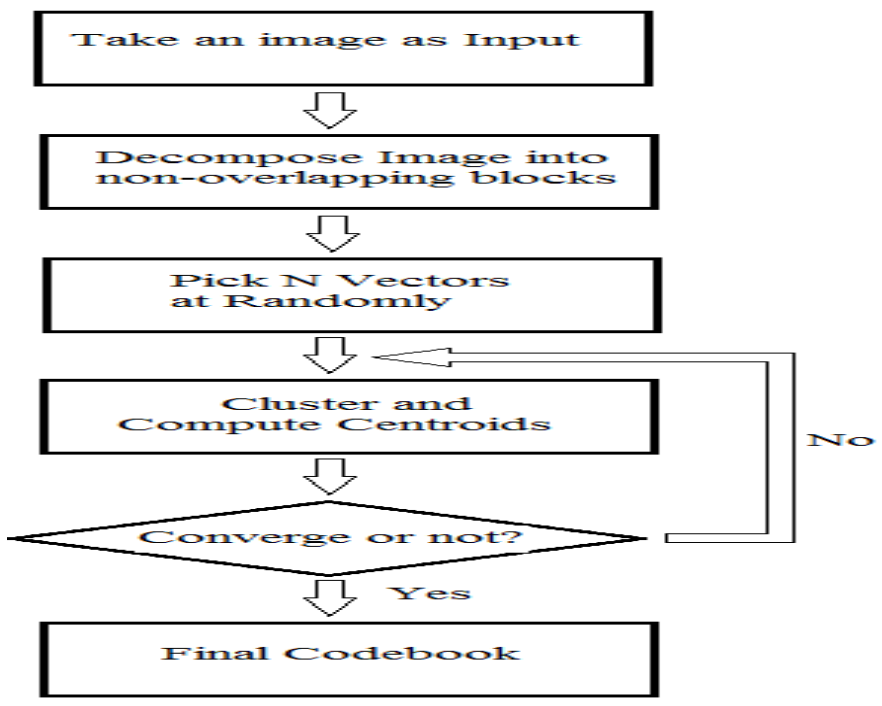

*Figure 1: The flowchart of LBG clustering algorithm*

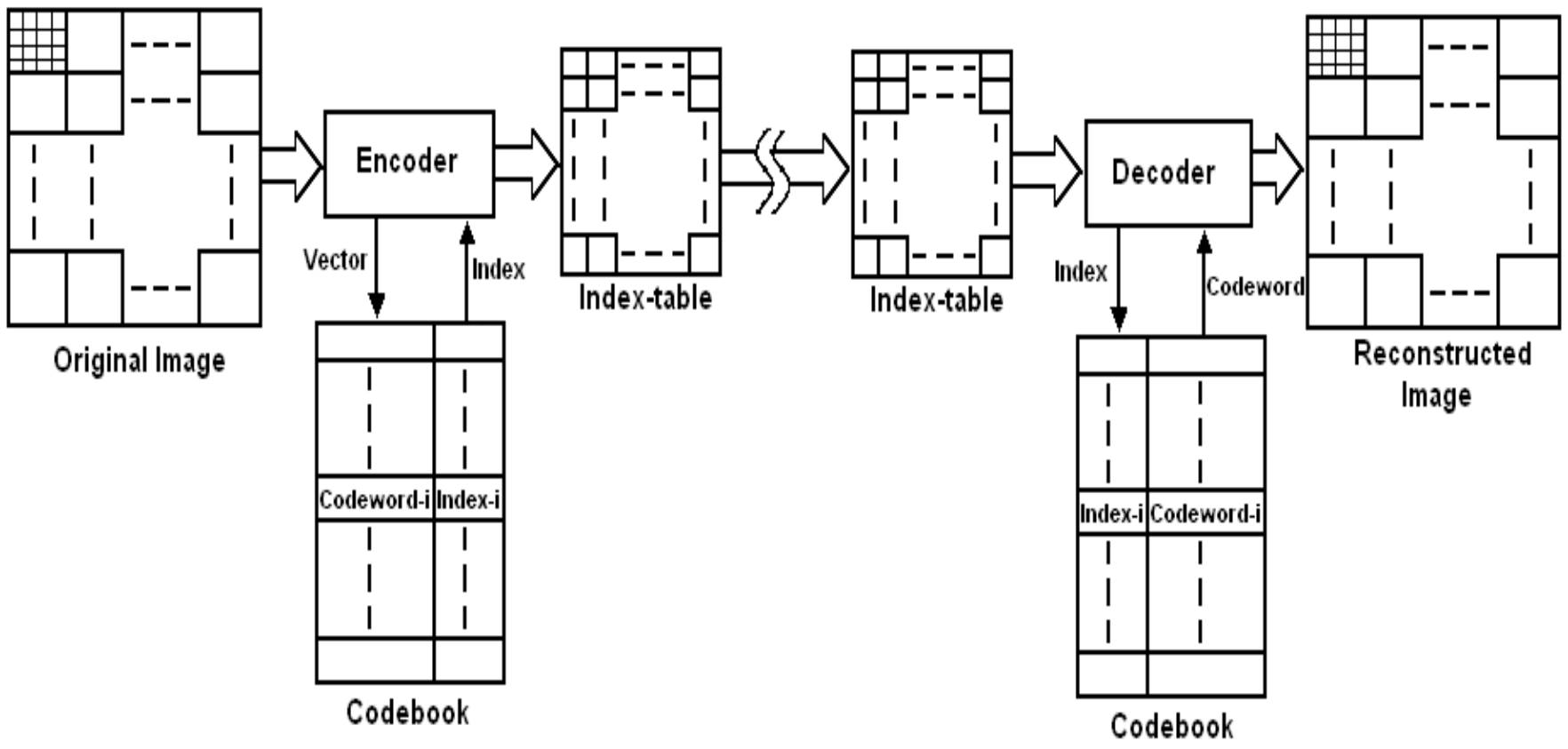

*Figure 2:* The Schematic diagram of VQ encoding-decoding process

## 3. THE PROPOSED TECHNIQUE

In this section, the proposed modified *LBG* algorithm using image pyramid is presented. The image pyramid offers a flexible, convenient multiresolution arrangement that mirrors the multiple scales of processing in the human visual system. It consists of a sequence of copies of an original image in which both sample density and resolution are decreased in regular steps. These reduced resolution levels of the pyramid are themselves obtained through a highly efficient iterative algorithm. The bottom of the pyramid represents the original image. After that, this is low pass-filtered and subsampled by a factor of two to obtain the next pyramid level. Further repetitions of the filter and subsample steps generate the remaining pyramid levels. An image pyramid for "Lena" image is shown in *Figure 3*.

In the proposed scheme, we have chosen the highest level reduced image. The highest level reduced image represents the coarse image of the original image. It also contains most of the represented pattern of the original image. So, in the proposed scheme, if we decompose this highest level coarse image in blocks, then collection of theses blocks are considered as pattern space and are called training vectors. Thus, in the proposed scheme, more priority is given to the vectors having more feature variations and this can be achieved by considering the highest level reduced image from the image pyramid. The flowchart of the modified LBG clustering algorithm is shown in Figure 4. The major steps of image compression and decompression are summarized as follows.

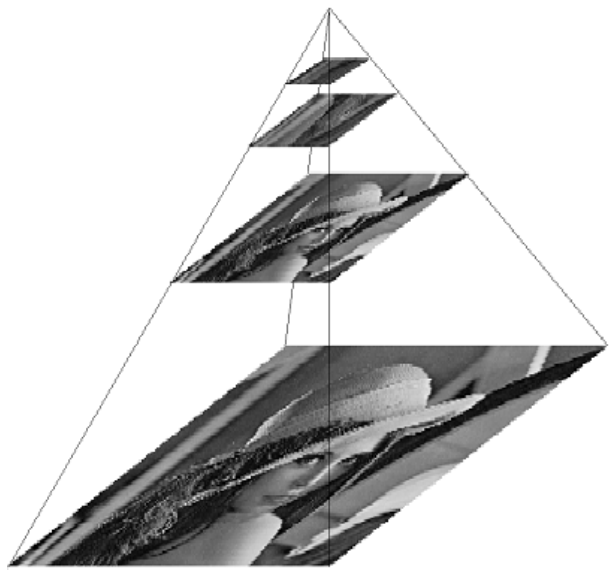

*Figure 3:* The image pyramid of the image Lena.

**Encoding:**

Step 1: Construct image pyramid of the input image and select the highest level reduced image.

Step 2: Decompose the obtained reduced image from *step 1* into non-overlapping blocks of size $n \times n$ pixels and convert each blocks into vectors.

Step 3: Collect all the vectors from *step 2* and use them as an initial codebook for VQ process.

Step 4: Improve the initial codebook into an improved one through several iterative process.

Step 5: Perform VQ encoding on the input image using the produced codebook and preserve the index-table.

Step 5: Store or transmit the codebook and the index-table as a compressed file of the input image.

**Decoding:**

Step 1: Perform VQ decoding using the index-table and the codebook to reconstruct the approximate image of the original image.

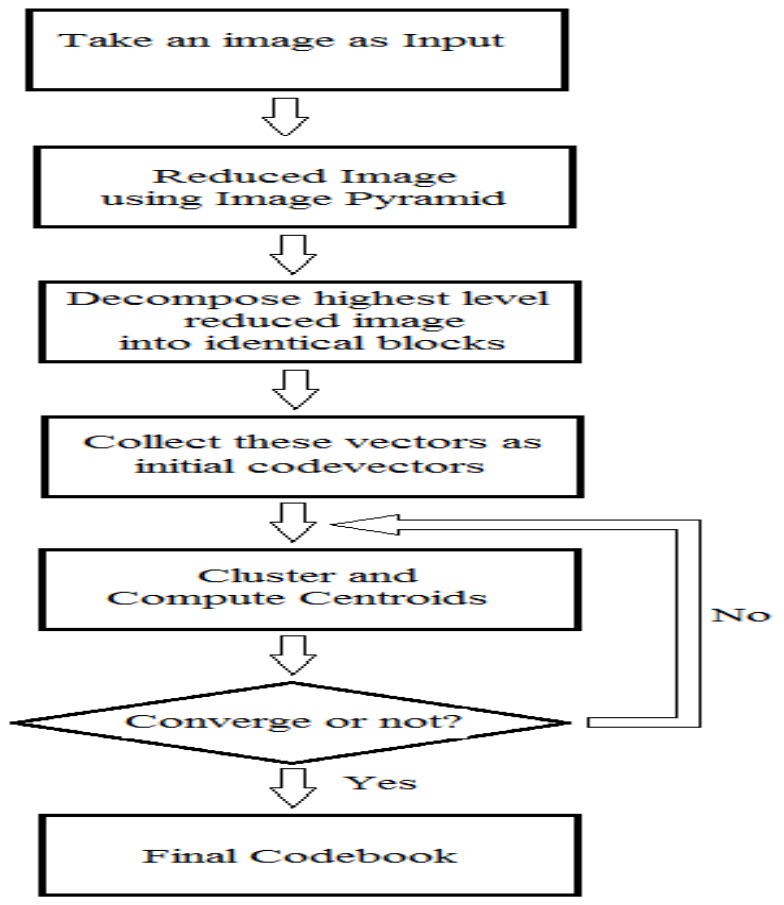

*Figure 4:* The flow chart of the modified LBG algorithm

## 4. EXPERIMENTAL RESULTS

In this section, the simulation results are presented to evaluate the performance of the proposed initialization method. We have carried out our proposed algorithm on a set of standard gray level images, but in this paper only the results of the most popular image 'Lena', 'Airplane', 'Girl' and 'Couple' of size 512×512 with 256 gray levels are considered. All original images are shown in *Figure 5*. The experimental results are presented into two parts: the first part is to evaluate the convergence speed of the final codebook by the proposed initialization method, and the second part is to compute the *PSNR* value of the encoded image with the intention of know if the resultant codebook is good representative or not. The details of result of the two parts are described in following subsections.

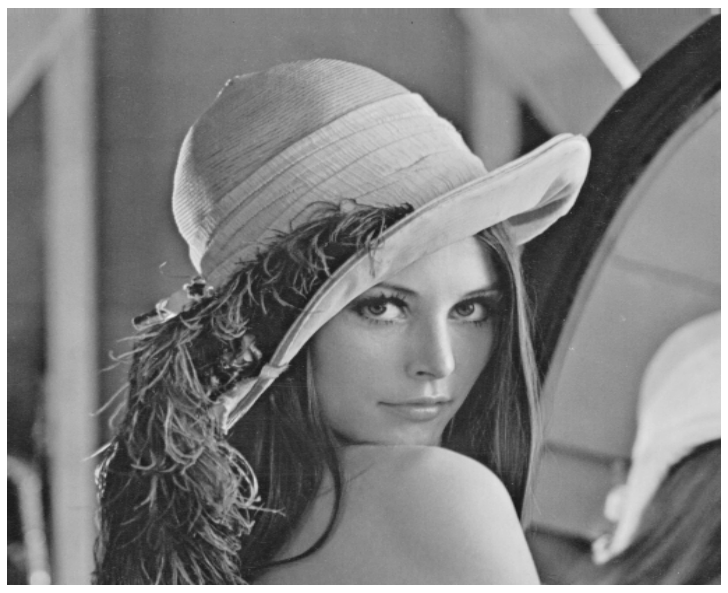
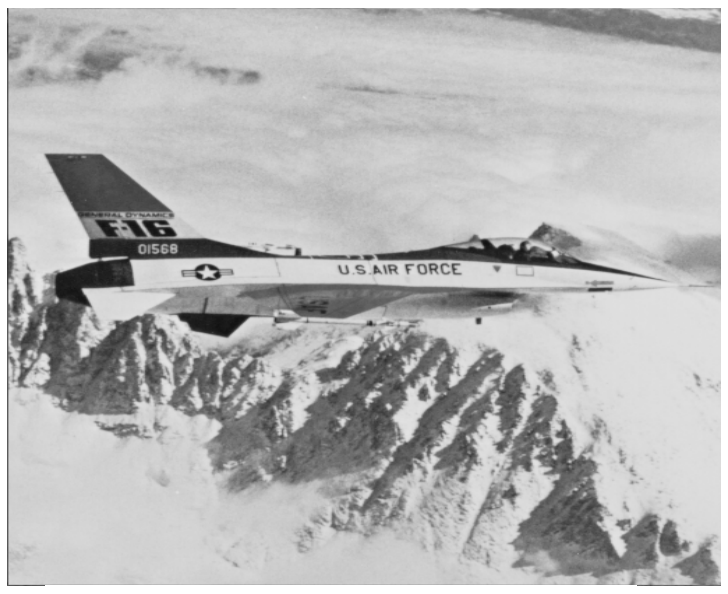

(a) Lena (b) Airplane

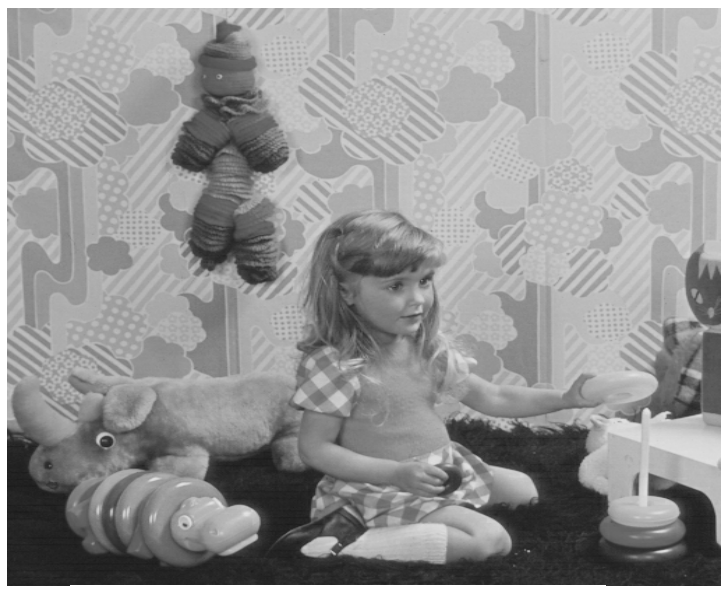
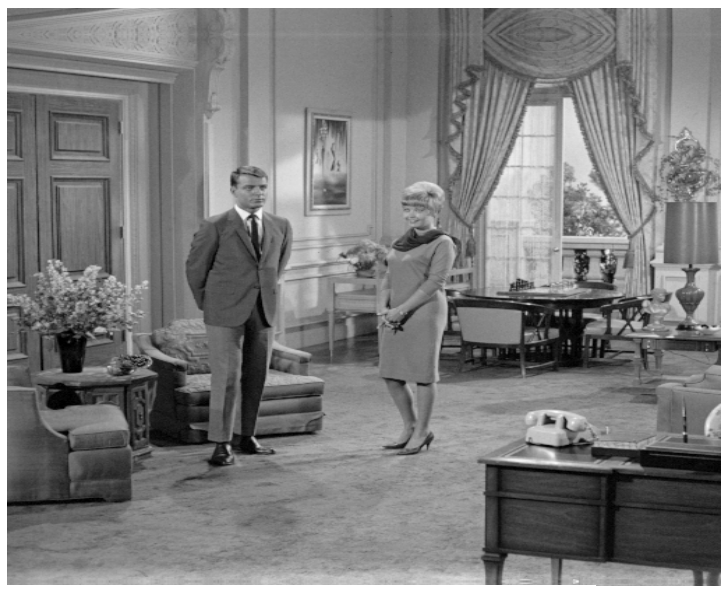

(c) Girl (d) Couple

*Figure 5:* The standard test images

### 4.1. Computation of the Convergence Rate of the Final Codebook

In this section, the required number of iteration for convergence of the final codebook is computed. Related computations have been performed using MATLAB 7.5.0 on the platform Intel Core2Duo 3.00 GHz, 4GB RAM, Microsoft Windows 7. In the proposed scheme, firstly we have taken a reduced size image from the image pyramid and subsequently the reduced image is decomposed into non-overlapping blocks. Later, the collection of all these vectors is used as an initial codebook. The LBG algorithm and the proposed algorithm have been used to generate codebook sized *128, 256, 512* and *1024* to test the speed of convergence. The experimental results are presented in *Table 1* where we have considered stopping threshold value, $\varepsilon = 0.001$. From the *Table 1*, it is shown that in most of the cases, the proposed scheme has successfully reduced the number of iterations compared to the LBG algorithm. So, the proposed scheme has improved the convergence rate as compared to the LBG algorithm.

### 4.2. Quality Measurement of the Encoded Images

In general, the measure of *PSNR* can be applied to evaluate the quality of the encoded images. *PSNR* measurement is adopted here to indicate the difference between the original image and its VQ decoded image. The *PSNR* values of the encoded images are listed in *Table 2*. From *Table 2*, it is noted that the proposed scheme produces predictable *PSNR* values and in some cases it produces better *PSNR* than the conventional *LBG* algorithm. So, based on the above experimental results, it is concluded that the proposed scheme outperforms than the *LBG* algorithm in terms of convergence speed for the training codebook and the *PSNR* performance for the encoded image quality. For both cases, the VQ decoded images (only for Lena) are presented in *Figure 6-7*.

Table 1: Performance Comparison in the number of iterations:

| Codebook Size | Block Size | Reduced Image Size for Modified LBG | Methods | Test Images (512 × 512 ) | | | |
|---|---|---|---|---|---|---|---|
| | | | | Lena | Airplane | Girl | Couple |
| 128 | 4×8 | 64×64 | Con. LBG | 9 | 16 | 7 | 6 |
| | | | Modified LBG | 7 | 12 | 4 | 5 |
| 128 | 8×4 | 64×64 | Con. LBG | 8 | 15 | 8 | 7 |
| | | | Modified LBG | 7 | 10 | 7 | 5 |
| 256 | 4×4 | 64×64 | Con. LBG | 10 | 14 | 11 | 9 |
| | | | Modified LBG | 8 | 11 | 8 | 6 |
| 256 | 8×8 | 64×64 | Con. LBG | 10 | 12 | 11 | 9 |
| | | | Modified LBG | 7 | 8 | 8 | 5 |
| 512 | 4×8 | 128×128 | Con. LBG | 11 | 14 | 14 | 11 |
| | | | Modified LBG | 9 | 10 | 9 | 7 |
| 512 | 8×4 | 128×128 | Con. LBG | 11 | 13 | 15 | 14 |
| | | | Modified LBG | 7 | 8 | 11 | 9 |
| 1024 | 4×4 | 128×128 | Con. LBG | 14 | 14 | 20 | 13 |
| | | | Modified LBG | 9 | 10 | 14 | 9 |
| 1024 | 8×8 | 128×128 | Con. LBG | 12 | 11 | 15 | 9 |
| | | | Modified LBG | 8 | 9 | 10 | 6 |

Table 2: Performance Comparison in Image Quality (PSNR)

| Codebook Size | Block Size | Reduced Image Size for Modified LBG | Methods | Test Images (512 × 512 ) | | | |
|---|---|---|---|---|---|---|---|
| | | | | Lena | Airplane | Girl | Couple |
| 128 | 4×8 | 64×64 | Con. LBG | 28.5966 | 27.4777 | 29.2377 | 27.1047 |
| | | | Modified LBG | 28.4320 | 27.2081 | 29.4290 | 26.8003 |
| 128 | 8×4 | 64×64 | Con. LBG | 29.0328 | 27.2106 | 29.3181 | 27.0125 |
| | | | Modified LBG | 29.0201 | 27.5767 | 28.8784 | 27.0181 |
| 256 | 4×4 | 64×64 | Con. LBG | 31.4367 | 28.5419 | 30.2956 | 27.8674 |
| | | | Modified LBG | 31.5946 | 28.2741 | 30.1090 | 28.0024 |
| 256 | 8×8 | 128×128 | Con. LBG | 27.8901 | 24.6388 | 26.7625 | 24.8225 |
| | | | Modified LBG | 27.7713 | 24.7752 | 26.4360 | 24.7217 |
| 512 | 4×8 | 128×128 | Con. LBG | 29.9681 | 29.4710 | 31.1701 | 28.4714 |
| | | | Modified LBG | 29.9213 | 29.5001 | 30.9802 | 28.5126 |
| 512 | 8×4 | 128×128 | Con. LBG | 30.5649 | 29.3145 | 31.0714 | 28.5620 |
| | | | Modified LBG | 30.2232 | 29.2916 | 31.0316 | 28.6012 |
| 1024 | 4×4 | 128×128 | Con. LBG | 33.1205 | 30.5173 | 32.0576 | 29.4459 |
| | | | Modified LBG | 33.2509 | 30.4812 | 31.9247 | 29.5912 |
| 1024 | 8×8 | 256×256 | Con. LBG | 28.5832 | 27.0164 | 28.4240 | 27.3339 |
| | | | Modified LBG | 28.7738 | 26.8531 | 28.3123 | 27.2831 |

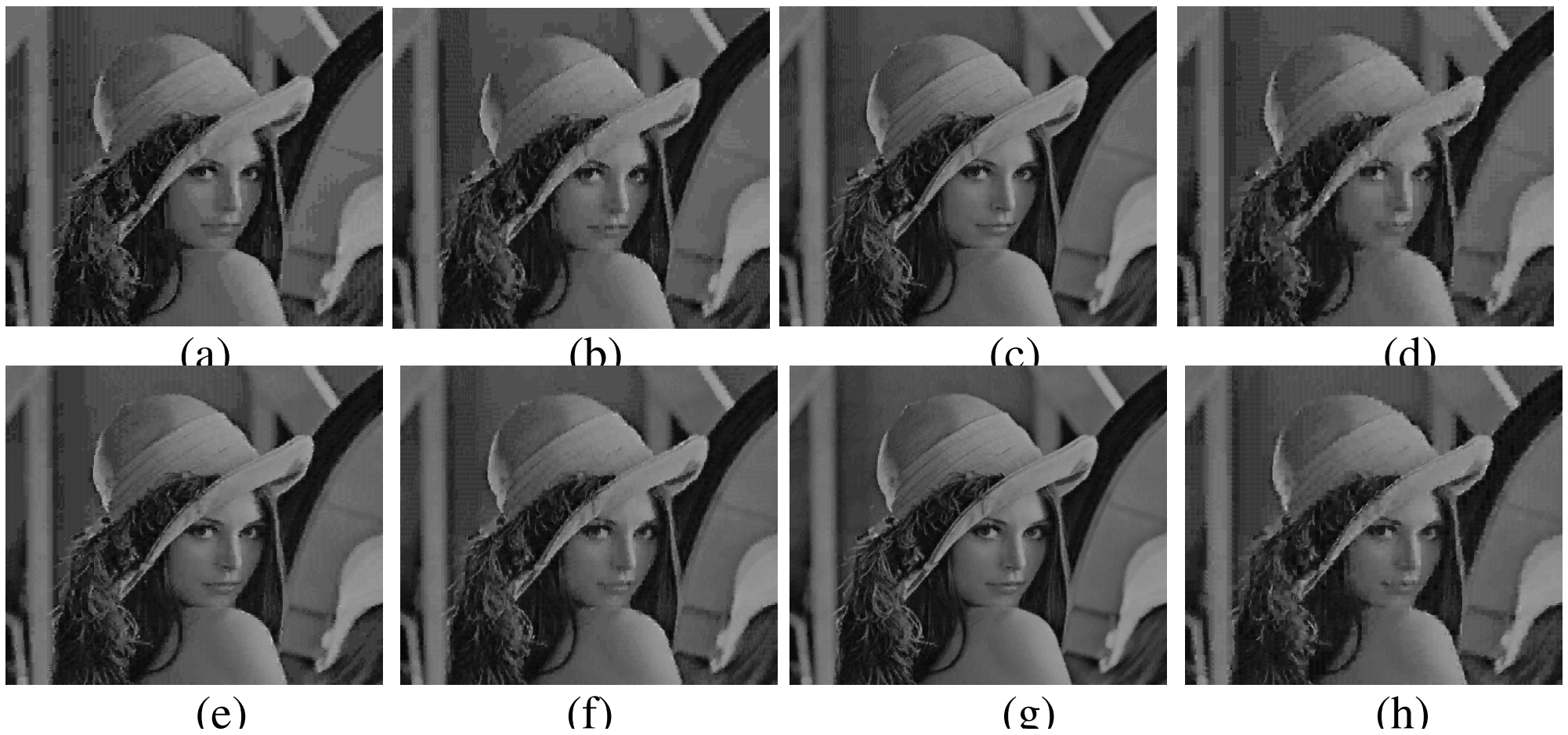

*Figure 6:* Reconstructed Image from LBG based trained codebook when (a) the codebook size 128 and codeword size 4×8; (a) the codebook size 128 and codeword size 8×4; (a) the codebook size 256 and codeword size 4×4; (a) the codebook size 256 and codeword size 8×8; (a) the codebook size 512 and codeword size 4×8; (a) the codebook size 512 and codeword size 8×4; (a) the codebook size 1024 and codeword size 4×4; (a) the codebook size 1024 and codeword size 8×8;

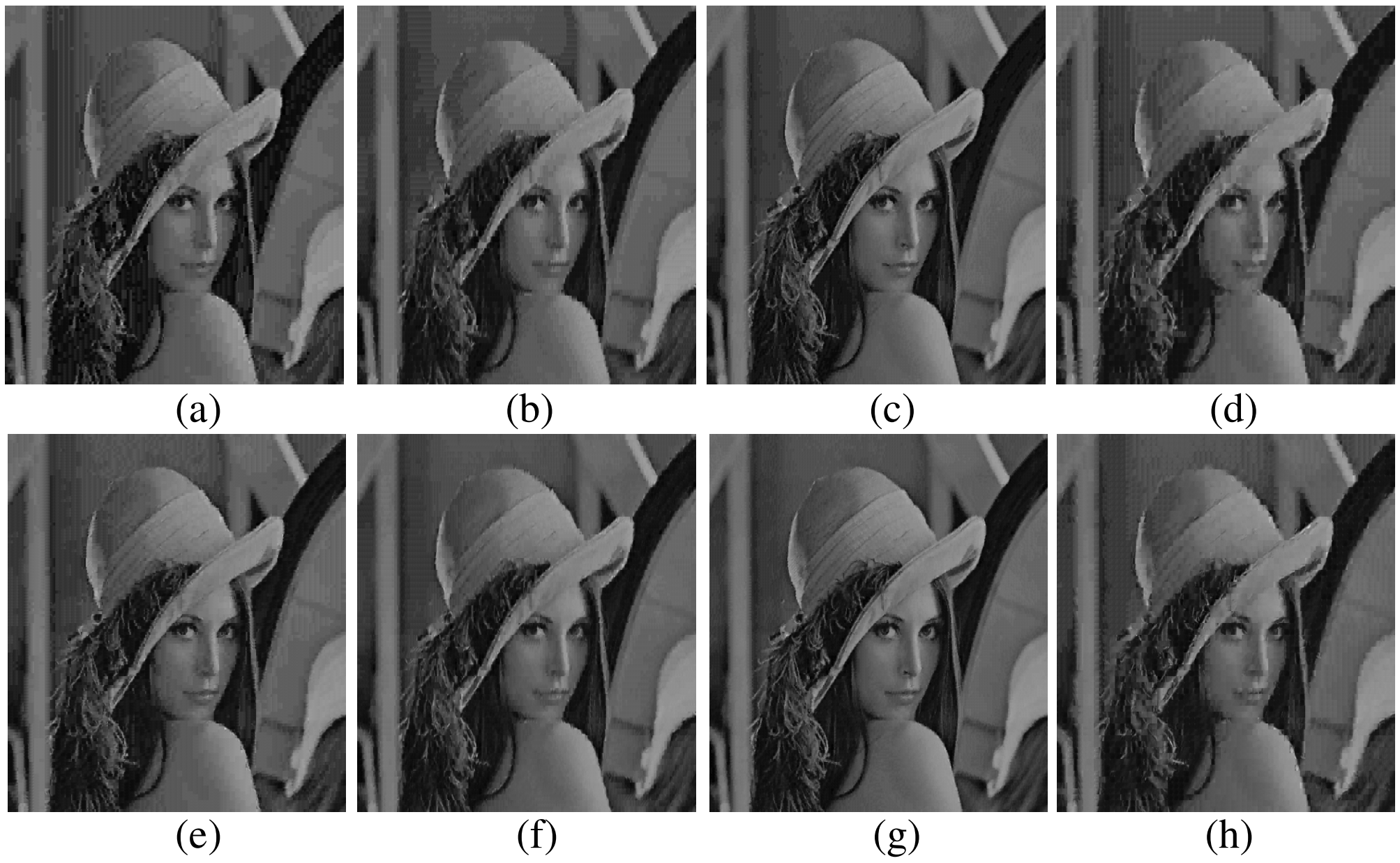

*Figure 7:* Reconstructed Image from Modified LBG based trained codebook when (a) the codebook size 128 and codeword size 4×8; (a) the codebook size 128 and codeword size 8×4; (a) the codebook size 256 and codeword size 4×4; (a) the codebook size 256 and codeword size 8×8; (a) the codebook size 512 and codeword size 4×8; (a) the codebook size 512 and codeword size 8×4; (a) the codebook size 1024 and codeword size 4×4; (a) the codebook size 1024 and codeword size 8×8;

## 5. CONCLUSIONS

A simple and efficient codebook initialization technique for the generation of a VQ codebook has been proposed in this paper. The initial good codebook is acquired from the highest level reduced image from the image pyramid. Finally, the resultant initial codebook is iteratively improved to become a final codebook. According to the experimental results, it is concluded that the proposed scheme outperforms the *LBG* algorithm in terms of convergence speed for the training codebook and the *PSNR* performance for the encoded image quality.